\documentclass{article}

\usepackage[nonatbib, final]{neurips_2021}

\usepackage[utf8]{inputenc} 
\usepackage[T1]{fontenc}    
\usepackage{hyperref}       
\usepackage{url}            
\usepackage{booktabs}       
\usepackage{amsfonts}       
\usepackage{nicefrac}       
\usepackage{microtype}      
\usepackage{xcolor}         
\usepackage{amsmath}

\usepackage{caption}
\usepackage{subcaption}
\usepackage{graphicx}
\usepackage{multirow}

\DeclareMathOperator*{\argmin}{arg\,min}

\title{Decomposing Representations for \\Deterministic Uncertainty Estimation}

\author{%
  Haiwen Huang  \qquad Joost van Amersfoort \qquad Yarin Gal \\
  OATML Group \\
  Department of Computer Science \\
  University of Oxford \\
  Oxford, United Kingdom \\
  \texttt{\{haiwen.huang2, joost.van.amersfoort, yarin\}@cs.ox.ac.uk} \\
}

\begin{document}

\maketitle


\section{Introduction}

Uncertainty estimation is a key component in any deployed machine learning system: when the system encounters unfamiliar or noisy input, it needs to refrain from making a decision and instead defer to an expert.
There has been great progress in uncertainty estimation in recent years, particularly in a sub-field called deterministic uncertainty estimation, where only a single forward pass is necessary to accurately estimate uncertainty \cite{mahalanobis,DDU,amersfoort2021improving}.

One way to evaluate uncertainty estimation is using ``out-of-distribution'' (OoD) detection, where it is expected that data points far away from the training distribution have higher uncertainty than points similar to the training distribution.
A concrete way to evaluate this is by distinguishing between two data distributions, one used for training and one never seen during training, using uncertainty.
Recently a challenging version of this benchmark was introduced: distinguishing CIFAR-10 from CIFAR-100 and vice versa.
This was called ``near'' OoD detection to distinguish from ``far'' OoD detection~\cite{contrastiveood}, for example in CIFAR-10 vs SVHN.
Near OoD detection is significantly more difficult as the near OoD data are more semantically similar to the training data. For example, CIFAR-10 and CIFAR-100 both come from the 80-million tiny images dataset, but contain no class overlap~\cite{cifar10}.

It was shown that in the near OoD setting, standard methods such as Mahalanobis distance~\cite{mahalanobis} and DDU~\cite{DDU} do not perform well~\cite{ren_simple_2021}.
We show that the proposed fix in~\cite{ren_simple_2021} does not work well in the far OoD setting, and propose a simple solution that does work.
We identify two types of features: discriminative (class-relevant) and non-discriminative (class-irrelevant) features.
Intuitively, one can consider the discriminative features  sufficient for distinguishing near OoD data, while non-discriminative features are used for far OoD detection.
In Table~\ref{pre-logit}, we show that when using a representation that is a mix of these features, uncertainty estimators based on feature density that perform well on one OoD detection benchmark have poor performance on the other.

This is because when the OoD data mainly differ in one specific type of representations, the other type can still influence the uncertainty.
To solve this, we propose to decompose the learned representations into two proposed types of representations and integrate the uncertainties estimated on them separately.
Through experiments, we demonstrate that we can greatly improve the performance and the interpretability of the uncertainty estimation.

\section{Motivating Observations}
First, we use Mahalanobis distance for OoD detection~\cite{mahalanobis} and its variants~\cite{DDU, contrastiveood, ren_simple_2021} to demonstrate the influence of class-relevance on uncertainty estimation. Since our analysis is focused on the feature maps, it can be extended to any other representation-based uncertainty estimation methods.

Given a pretrained feature extractor $f: \mathbb{R}^n \rightarrow \mathbb{R}^m$ and dataset $\{x_i, y_i\}_{i=1}^N$ where input $x_i \in \mathbb{R}^n$, and label $y_i \in \{1, \cdots, C\}$, the Mahalanobis distance for OoD detection first models the distribution of the features as a class-conditional multivariate Gaussian distribution with a single covariance matrix shared across classes.
That is, for each class $c$, it models the distribution of features as $p(z|y=c) \sim \mathcal{N}(\boldsymbol{\hat{\mu}}_c, \Sigma)$. It then calculates class-wise Mahalonobis distance (i.e. the same quantity as in the exponent of a multi-variate Gaussian distribution)  of a feature representation $z = f(x)$ and uses the maximum across classes as the distance:
\begin{equation}
    M(x) = \max_c - (f(x) - \boldsymbol{\hat{\mu}}_c)^T \hat{\Sigma}^{-1} (f(x) - \boldsymbol{\hat{\mu}}_c),
    \label{maha}
\end{equation}
where 
$ \hat{\Sigma} = \frac{1}{N} \sum_{c=1}^{C} \sum_{i: y_{i}=c}\left(f(\boldsymbol{x}_{i})-\boldsymbol{\hat{\mu}}_{c}\right)\left(f(\boldsymbol{x}_{i})-\boldsymbol{\hat{\mu}}_{c}\right)^{T},\; \boldsymbol{\hat{\mu}}_{c} = \sum_{i: y_{i}=c}f(\boldsymbol{x}_{i}).$ 
Recently, many works~\cite{DDU, contrastiveood} model the feature distribution using a \textit{class-dependent} covariance matrix and found it beneficial.
In this case, the uncertainty metric is $M(x)=\max_c [-\left(f(x)-\boldsymbol{\hat{\mu}}_{c}\right)^{T} \boldsymbol{\hat{\Sigma}}_{c}^{-1}\left(f(x)-\boldsymbol{\hat{\mu}}_{c}\right)-\log ((2 \pi)^{m} \operatorname{det} \mathbf{\hat{\Sigma}}_{c})]$ where $ \mathbf{\hat{\Sigma}}_c = \frac{1}{|\{i: y_{i}=c\}|} \sum_{i: y_{i}=c}(f(\boldsymbol{x}_{i})-\boldsymbol{\hat{\mu}}_{c})(f(\boldsymbol{x}_{i})-\boldsymbol{\hat{\mu}}_{c})^{T}.$

Variants of Mahalanobis distance were proposed to improve on \textit{some specific tasks}. 
Notably, marginal Mahalanobis distance (\textit{Marginal Maha})~\cite{kamoi_why_2020} calculates the mean and covariance without using class-relevant information. Specifically, in Equation~\eqref{maha}, 
$\Sigma = \frac{1}{N} \sum_{i=1}^{N}\left(f(\boldsymbol{x}_{i})-\boldsymbol{\mu}\right)\left(f(\boldsymbol{x}_{i})-\boldsymbol{\mu}\right)^{T},\; \boldsymbol{\mu}_c = \boldsymbol{\mu} = \frac{1}{N}\sum_{i=1}^{N}f(\boldsymbol{x}_{i}).$
Another variant is called Relative Mahalanobis distance (\textit{Relative Maha})~\cite{ren_simple_2021} $ M_{rel}(x) = M(x) - M_{marg}(x)$ where $M_{rel}$ and $M_{marg}$ refers to \textit{Relative Maha} and \textit{Marginal Maha}, respectively. 
It aims to mitigate the influence of the class-irrelevant information.
As we can see from the definitions, the key difference among the three variants is how the class-relevant information is (not) used. 
In terms of the degree of class-relevance: \textit{Relative Maha} $>$ \textit{Maha} $>$ \textit{Marginal Maha}.

To empirically compare these three methods, we use two OoD detection benchmarks: CIFAR10~\cite{cifar10} vs. SVHN~\cite{svhn} and CIFAR10 vs. CIFAR100. 
The OoD detection results can be found in Table~\ref{pre-logit}.
We can see that on CIFAR10 vs. SVHN, all the methods can perform relatively well with \textit{Maha} being the top. While on CIFAR10 vs. CIFAR100, there is a sharp difference among the methods: \textit{Relative Maha} shows significantly better performance than others while \textit{Marginal Maha} almost completely mixes the two datasets (AUROC=50\% means random guess). 
Note that although the compared methods are well-known, none of them has been tested \textit{simultaneously} on both benchmarks. Our experiments demonstrate that \textit{none of these methods can beat every other method on both benchmarks}.


\begin{table}
\centering
\caption{AUROC (\%) comparison of variants of Mahalanobis distance calculated on features at penultimate layer (pre-logit).}
\vspace{6pt}
\resizebox{0.7\columnwidth}{!}{%
\begin{tabular}{lcc}
\toprule
    Method        & CIFAR10 vs SVHN & CIFAR10 vs CIFAR100 \\
\midrule
    Maha~\cite{mahalanobis}     & \textbf{97.96}           & 75.75               \\
    Marginal Maha~\cite{kamoi2020mahalanobis} & 96.07         & 60.10               \\
    Relative Maha~\cite{ren_simple_2021} & 95.43           & \textbf{90.97}               \\
\bottomrule
\end{tabular}
}
\label{pre-logit}
\end{table}



\section{Decomposing Representations}\label{experiment}
Considering the difference among the three compared methods in the last section, we hypothesize that the differences in OoD detection performances can be attributed to the different \textit{degrees of class relevance}.
Therefore, we propose decomposing the discriminative and non-discriminative information from the learned representations. 
This allows us to estimate uncertainty on each type of representations separately and provides a new perspective to interpret the uncertainty estimation.

Formally, an ideal decomposition  of the features $z$ into the discriminative part $z_d$ and non-discriminative part $z_n$, i.e., $z=(z_d; z_n)$, should have the following properties:
\begin{itemize}
    \item $I(z_d; z_n | y) = 0$ (the two decomposed parts should be conditionally independent given the labels $y$);
    \item $I(z_n; y)=0$ ($z_n$ is non-discriminative, i.e., independent from the labels $y$);
    \item $z_d = \argmin_S I(z_d; z)$ where $S = \{z_d \mid I(z_d;y) = I(z;y)\}$
    ($z_d$ includes and only includes all the discriminative information in $z$).
\end{itemize}

To perform the decomposition, a simple way is to use Principal Component Analysis (PCA) to transform the features and then take the first $d$ principal components (PCs) as $z_d$ and the remaining PCs as $z_n$. However, since PCA is a linear method and does not use any label information, there is no guarantee that PCA can achieve the ideal decomposition. Alternatively, we can use the independent cross-entropy (iCE) loss proposed by~\cite{jacobsen2018excessive}.
Specifically, given a fixed pretrained feature extractor $f: \mathbb{R}^M \rightarrow \mathbb{R}^D$ which maps from inputs to representations, we first perform a transformation of the features $z=f(x)$ using an invertible function $F: \mathbb{R}^D \rightarrow \mathbb{R}^D$. The invertibility is to make sure we keep all the information of the features after the transformation. Our aim then is to extract discriminative features $z_d$ into the first $d$ dimensions of $F(z)$ and non-discriminative features $z_n$ into the other dimensions. 
As a convenient choice, we set $d=C$, i.e., the number of classes of the training dataset. $z_d$ then serves as logits for the classification. 
We then uses another linear map $D: \mathbb{R}^{D-d} \rightarrow \mathbb{R}^C$ to map $z_n$ to a $C-$dimensional logits. In this way, we can now use the iCE loss to encourage the decomposition of discriminative and non-discriminative features. The iCE loss is defined as: 
\begin{equation}
    \min _{\theta} \max _{\phi} \mathcal{L}_{i C E}\left(\theta, \phi\right)=
    \sum_{i=1}^{C}-y_{i} \log [\text{softmax} [F_{\theta}(z)]]_{i}
    +
    \sum_{i=1}^{D-C} y_{i} \log [\text{softmax} [D_{\phi}\left(F_{\theta}(z)_{[C+1:D]}\right)]]_{i}.
\end{equation}
We point out that for the second loss term, the minimization is concerning a lower bound on $I(y; z_n)$, while the maximization aims to tighten the bound~\cite{jacobsen2018excessive}. Therefore, the iCE loss aims to maximize $I(y; z_s)$ (first term) and minimize $I(y; z_n)$ (second term) at the same time to achieve the second and third properties in our definition of decomposition (note the minimization of $I(z;z_d)$ is also achieved during optimization~\cite{tishby2000information}).
The first independence property is not explicitly enforced in our implementation, but can still be achieved through the optimization. We discuss in the appendix~\ref{more_decomposition}.

When the decomposition is finished, we can get a new uncertainty estimate based on the uncertainty calculated on the decomposed features. Because ideally, $\log p(z)  = \log p(z_d; z_n) = \log p(z_d) + \log p(z_n)$ (since $I(z_d;z_n)=0$), the new uncertainty estimate can simply  be $M'(z) := 1/2 M (z_d) + 1/2 M (z_n)$, where $M$ is an uncertainty estimator on the features, e.g., \textit{Maha}.
We can also design a \textit{KL-based dataset distance metric} based on the decomposition: $d (D_{in}, D_{out}) = \mathcal{D}_{KL}[p_{in}(M(z))\|p_{out}(M(z))]$, where $p_{in}$ and $p_{out}$ are the distributions of the OoD scores (e.g. \textit{Maha} distance) on $\mathcal{D}_{in}$ and $\mathcal{D}_{out}$ respectively. Depending on the features we use (i.e., $z_d$ or $z_n$), we can define $d_{dis}$ and $d_{nondis}$ correspondingly. Details can be seen in the appendix~\ref{distance}.






\section{Experiments}

\begin{table}
    \centering
    \caption{AUROC (\%) comparison of different methods calculated on features at penultimate layer (pre-logit). $D_{in}$ = CIFAR10.}
    \vspace{6pt}
    \resizebox{\columnwidth}{!}{%
    \begin{tabular}{llll}
    \toprule
    \multirow{5}{*}{\textbf{Method}} & \multicolumn{3}{c}{\textbf{Datasets}}  \\
                    & $D_{out}$ = SVHN & $D_{out}$ =CIFAR100 & $D_{out}$ = SVHN $\cup$ CIFAR100 \\
                 &  $d_{dis} = 3.75$  &  $d_{dis} =4.55$ & $d_{dis}=2.08$   \\ 
                    &  $d_{nondis} =4.09$ & $d_{nondis} =1.15$ & $d_{nondis}=1.73$  \\ 
    \midrule
        Maha           & 97.96           &  75.75  &  86.86     \\
        Marginal Maha &  96.07       & 60.10   & 78.08 \\
        Relative Maha &  95.43       & \textbf{90.97}      &    93.20 \\
        \midrule
        PCA top score & 94.59 & 90.20 & 92.39\\
        PCA bottom score & 98.23 & 83.77 & 91.00 \\
        PCA top score + bottom score & 97.95 & 89.90 & 93.92 \\
        Discriminative score & 96.14 & 90.46 & 93.30 \\
        Non-discriminative score & 98.08 & 76.14 & 87.11 \\
        Dis score + Non-dis score & \textbf{98.41} & 89.95 & \textbf{94.18} \\
    \bottomrule
    \end{tabular}
    }
    \label{main}
\end{table}

\subsubsection*{Main Result}
In Table~\ref{main}, we show the comparison of different methods on the three OoD benchmarks.
The scoring function we use is \textit{Maha}.
From Table~\ref{main}, we can see that: 
\begin{enumerate}
    \item Our dataset distance metric is a good indicator of the OoD types and the AUROC differences.
    SVHN is far from CIFAR10 in both features but $d_{nondis}$ is slightly larger. So while either type of features can achieve high AUROC for detecting SVHN, non-discriminative features are slightly better. 
    For CIFAR100, $d_{dis}$ is much larger than $d_{nondis}$. So when using discriminative features to detect CIFAR100, we can see considerable improvement.
    \item Discriminative score performs best on discriminative features, and non-discriminative score performs best on non-discriminative features. Moreover, the sum of the two scores combines the best of both features, yielding close to best performances on both benchmarks and best performance on the union benchmark.
    \item PCA is a strong baseline for the decomposition. 
    Note that this conclusion can change when we are not using the feature extractor of a  discriminatively trained neural network.
\end{enumerate}

\begin{figure}
    \centering
    \begin{subfigure}{0.48\textwidth}
         \centering
         \includegraphics[width=\textwidth]{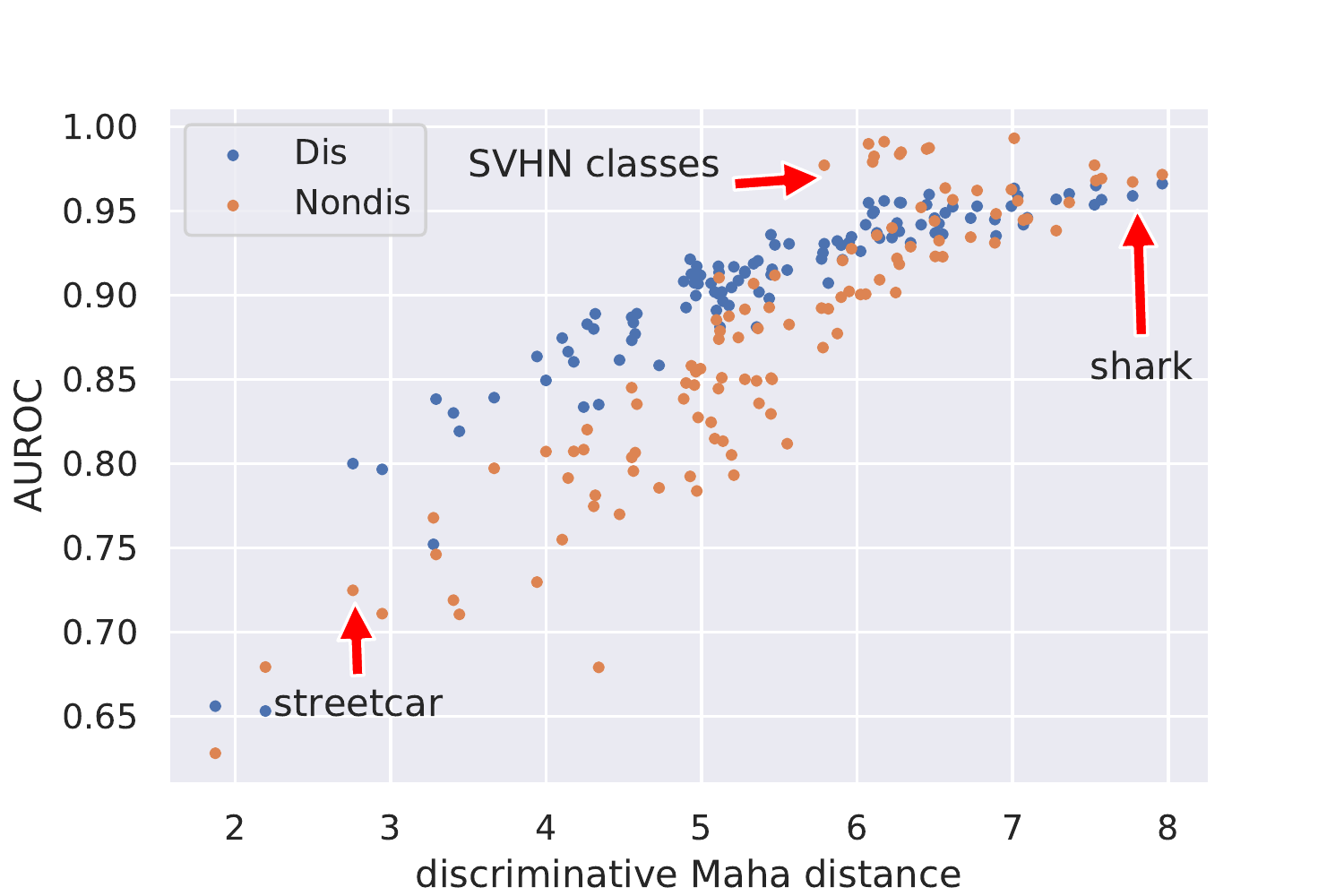}
         \caption{}
         \label{dis_auroc}
    \end{subfigure}
    \begin{subfigure}{0.48\textwidth}
         \centering
         \includegraphics[width=\textwidth]{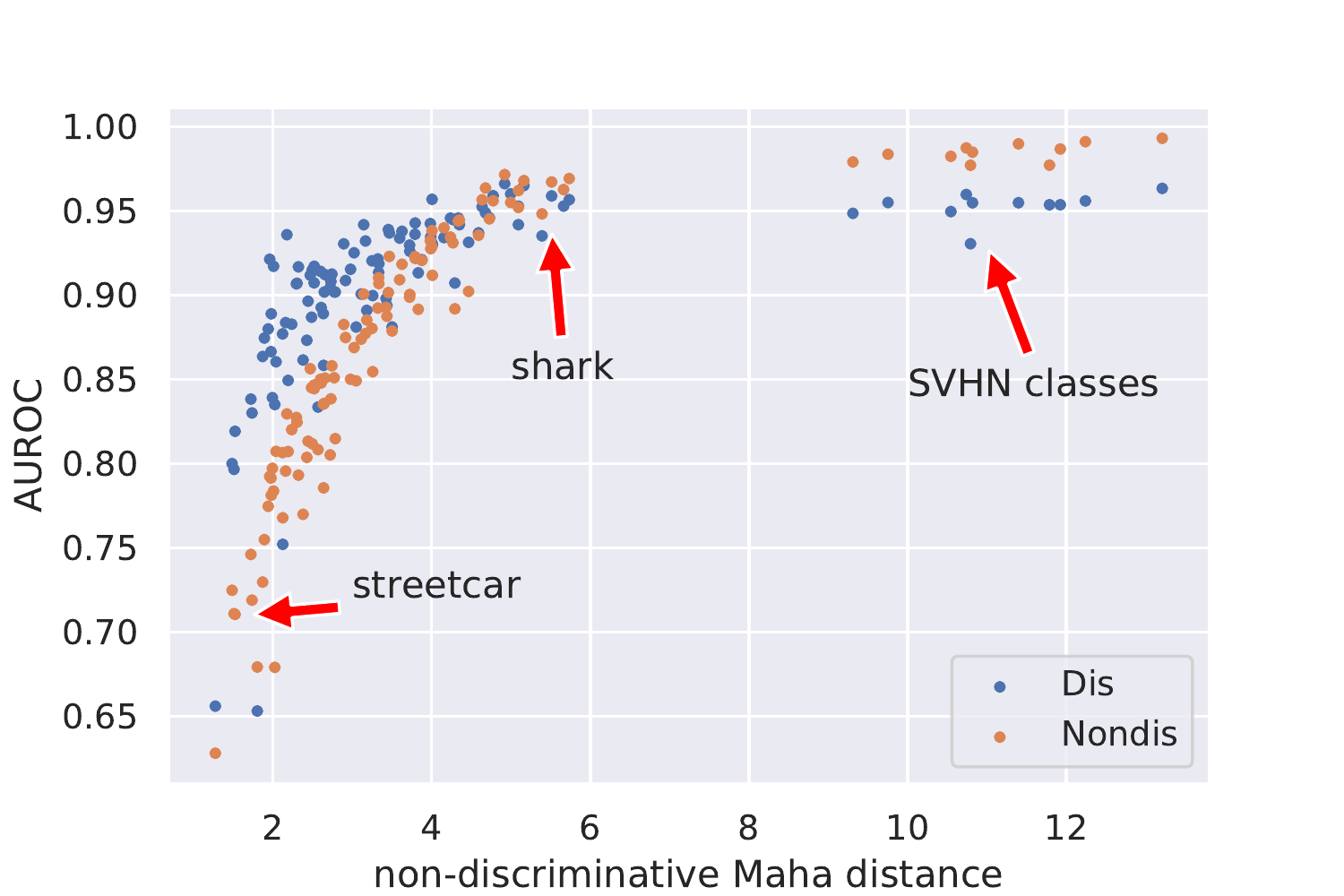}
         \caption{}
         \label{nondis_auroc}
    \end{subfigure}
    \caption{AUROCs of OoD detection using discriminative and non-discriminative features (y-axis) against the averaged Maha distances on the two features for each class (x-axis). Every two dots with the same x-axis coordinate belong to the same class in either SVHN or CIFAR100.}
    \label{scatter}
\end{figure}



\subsection{Interpreting Uncertainty}
The dataset distance under different features also provides us with a tool to interpret the uncertainty estimates. 
In Figure~\ref{scatter}, we conduct a class-wise analysis for a fine-grained understanding of the OoD classes. Specifically, we calculate the AUROC using discriminative and non-discriminative scores to detect the 110 classes of SVHN $\cup$ CIFAR100 from CIFAR10 (one class at a time). This gives us two AUROC scores for each class, \textit{dis AUROC} and \textit{nondis AUROC}. We then plot them against the dataset distance between CIFAR10 and each class. In this way, we can interpret the OoD detection performances (y-axis) by looking at dataset distances (x-axis). 

We highlight three types of OoD: (1) Classes with much higher non-discriminative distances than discriminative ones, e.g., \textit{SVHN classes}. For these classes, non-discriminative features are more suitable for OoD detection. (2) Classes with small discriminative distances, e.g., \textit{streetcar}. The small discriminative distances are usually due to similar categories in CIFAR10, e.g., automobile and truck for the streetcar. These classes typically also have small non-discriminative distances. Usually, discriminative distances are still larger, so more suitable for OoD detection. (3) Classes with large discriminative distances, e.g., \textit{shark}. These classes typically also have large non-discriminative distances. The OoD performances using the two kinds of distances are usually similar.

\section{Discussion}

In this work, we showed that current state-of-the-art uncertainty estimation methods could not consistently outperform other methods across different benchmarks. We solved this problem by decomposing the features. Specifically, we summed the OoD scores calculated separately on the discriminative and non-discriminative features and achieved consistently high performance across different types of benchmarks.
Our decomposition can also help us interpret the uncertainty by looking at the uncertainty estimates on the decomposed features.

\bibliography{reference}

\begin{thebibliography}{10}

\bibitem{mahalanobis}
Kimin Lee, Kibok Lee, Honglak Lee, and Jinwoo Shin.
\newblock A simple unified framework for detecting out-of-distribution samples
  and adversarial attacks.
\newblock In {\em Proceedings of the 32nd International Conference on Neural
  Information Processing Systems}, NIPS’18, page 7167–7177, Red Hook, NY,
  USA, 2018. Curran Associates Inc.

\bibitem{DDU}
Jishnu Mukhoti, Andreas Kirsch, Joost van Amersfoort, Philip H.~S. Torr, and
  Yarin Gal.
\newblock Deterministic neural networks with appropriate inductive biases
  capture epistemic and aleatoric uncertainty.
\newblock {\em CoRR}, abs/2102.11582, 2021.

\bibitem{amersfoort2021improving}
Joost van Amersfoort, Lewis Smith, Andrew Jesson, Oscar Key, and Yarin Gal.
\newblock Improving deterministic uncertainty estimation in deep learning for
  classification and regression, 2021.

\bibitem{contrastiveood}
Jim Winkens, Rudy Bunel, Abhijit~Guha Roy, Robert Stanforth, Vivek Natarajan,
  Joseph~R. Ledsam, Patricia MacWilliams, Pushmeet Kohli, Alan
  Karthikesalingam, Simon Kohl, A.~Taylan Cemgil, S.~M.~Ali Eslami, and Olaf
  Ronneberger.
\newblock Contrastive training for improved out-of-distribution detection.
\newblock {\em CoRR}, abs/2007.05566, 2020.

\bibitem{cifar10}
Alex Krizhevsky.
\newblock Learning multiple layers of features from tiny images.
\newblock Technical report, 2009.

\bibitem{ren_simple_2021}
Jie Ren, Stanislav Fort, Jeremiah Liu, Abhijit~Guha Roy, Shreyas Padhy, and
  Balaji Lakshminarayanan.
\newblock A {Simple} {Fix} to {Mahalanobis} {Distance} for {Improving}
  {Near}-{OOD} {Detection}.
\newblock {\em arXiv:2106.09022 [cs]}, June 2021.
\newblock arXiv: 2106.09022.

\bibitem{kamoi_why_2020}
Ryo Kamoi and Kei Kobayashi.
\newblock Why is the {Mahalanobis} {Distance} {Effective} for {Anomaly}
  {Detection}?
\newblock {\em arXiv:2003.00402 [cs, stat]}, April 2020.
\newblock arXiv: 2003.00402.

\bibitem{svhn}
Yuval Netzer, Tao Wang, Adam Coates, Alessandro Bissacco, Bo~Wu, and Andrew~Y.
  Ng.
\newblock Reading digits in natural images with unsupervised feature learning.
\newblock In {\em NIPS Workshop on Deep Learning and Unsupervised Feature
  Learning 2011}, 2011.

\bibitem{kamoi2020mahalanobis}
Ryo Kamoi and Kei Kobayashi.
\newblock Why is the {Mahalanobis} {Distance} {Effective} for {Anomaly}
  {Detection}?
\newblock {\em arXiv:2003.00402 [cs, stat]}, April 2020.
\newblock arXiv: 2003.00402.

\bibitem{jacobsen2018excessive}
Joern-Henrik Jacobsen, Jens Behrmann, Richard Zemel, and Matthias Bethge.
\newblock Excessive invariance causes adversarial vulnerability.
\newblock In {\em International Conference on Learning Representations}, 2019.

\bibitem{tishby2000information}
Naftali Tishby, Fernando~C Pereira, and William Bialek.
\newblock The information bottleneck method.
\newblock {\em arXiv preprint physics/0004057}, 2000.

\bibitem{krizhevsky2009learning}
Alex Krizhevsky, Geoffrey Hinton, et~al.
\newblock Learning multiple layers of features from tiny images.
\newblock {\em Tech Report}, 2009.

\bibitem{zagoruyko2016wide}
Sergey Zagoruyko and Nikos Komodakis.
\newblock Wide residual networks.
\newblock In {\em BMVC}, 2016.

\bibitem{behrmann_invertible_2019}
Jens Behrmann, Will Grathwohl, Ricky T.~Q. Chen, David Duvenaud, and
  Joern-Henrik Jacobsen.
\newblock Invertible {Residual} {Networks}.
\newblock In {\em International {Conference} on {Machine} {Learning}}, pages
  573--582. PMLR, May 2019.
\newblock ISSN: 2640-3498.

\bibitem{hsic}
Arthur Gretton, Kenji Fukumizu, Choon Teo, Le~Song, Bernhard Sch\"{o}lkopf, and
  Alex Smola.
\newblock A kernel statistical test of independence.
\newblock In J.~Platt, D.~Koller, Y.~Singer, and S.~Roweis, editors, {\em
  Advances in Neural Information Processing Systems}, volume~20. Curran
  Associates, Inc., 2008.

\bibitem{wang2009divergence}
Qing Wang, Sanjeev~R Kulkarni, and Sergio Verd{\'u}.
\newblock Divergence estimation for multidimensional densities via $ k
  $-nearest-neighbor distances.
\newblock {\em IEEE Transactions on Information Theory}, 55(5):2392--2405,
  2009.

\end{thebibliography}

\newpage

\appendix

\section{Additional Experiments}

\subsection{Toy experiments}

\begin{figure}
    \centering
    \begin{subfigure}{0.325\textwidth}
         \centering
         \includegraphics[width=\textwidth]{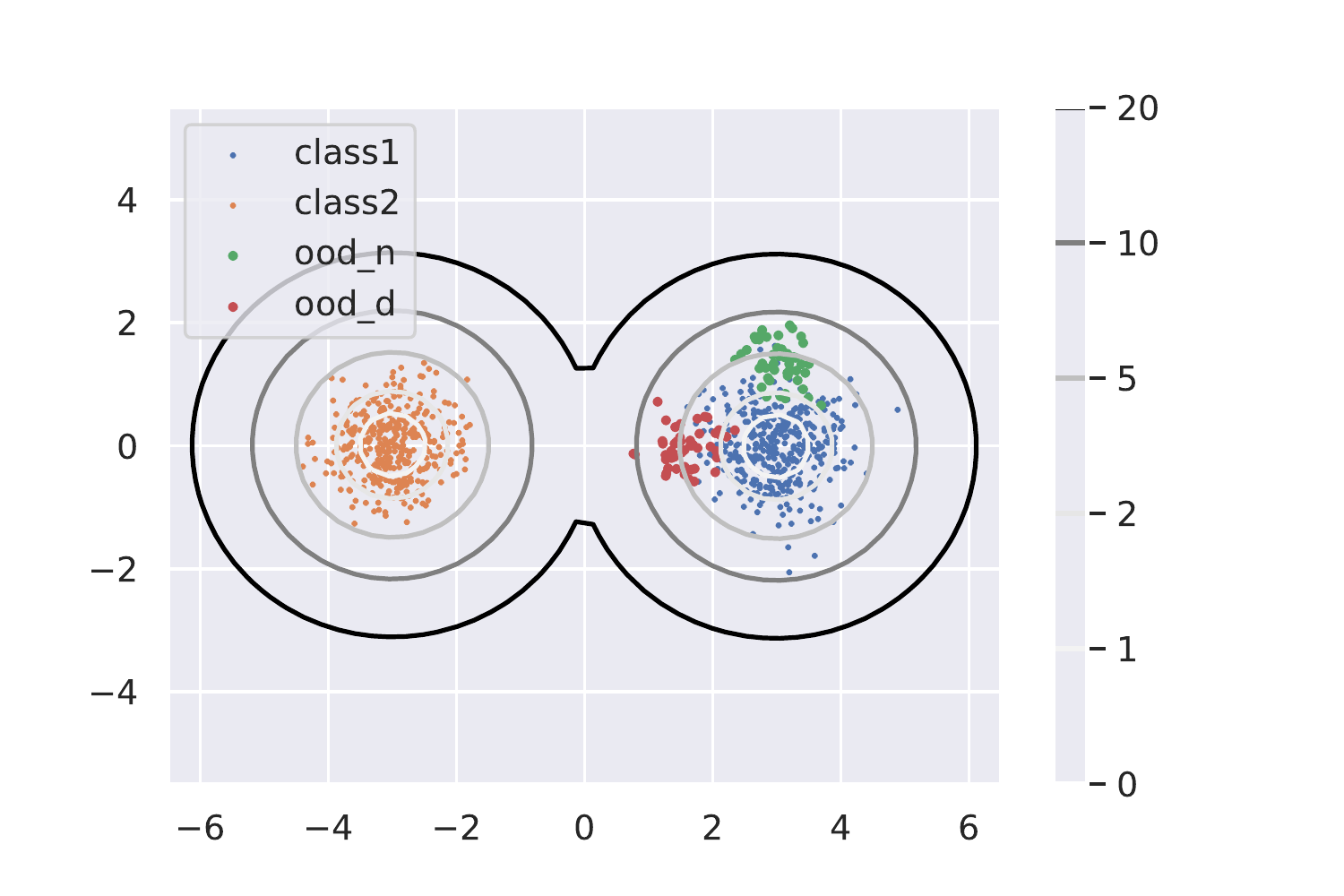}
          \caption{Mahalanobis distance on full features. \newline  AUROC$_{red}$ = 0.94, \newline  $\;$ AUROC$_{green}$=0.94.}
         \label{analytic}
    \end{subfigure}
    \begin{subfigure}{0.325\textwidth}
         \centering
         \includegraphics[width=\textwidth]{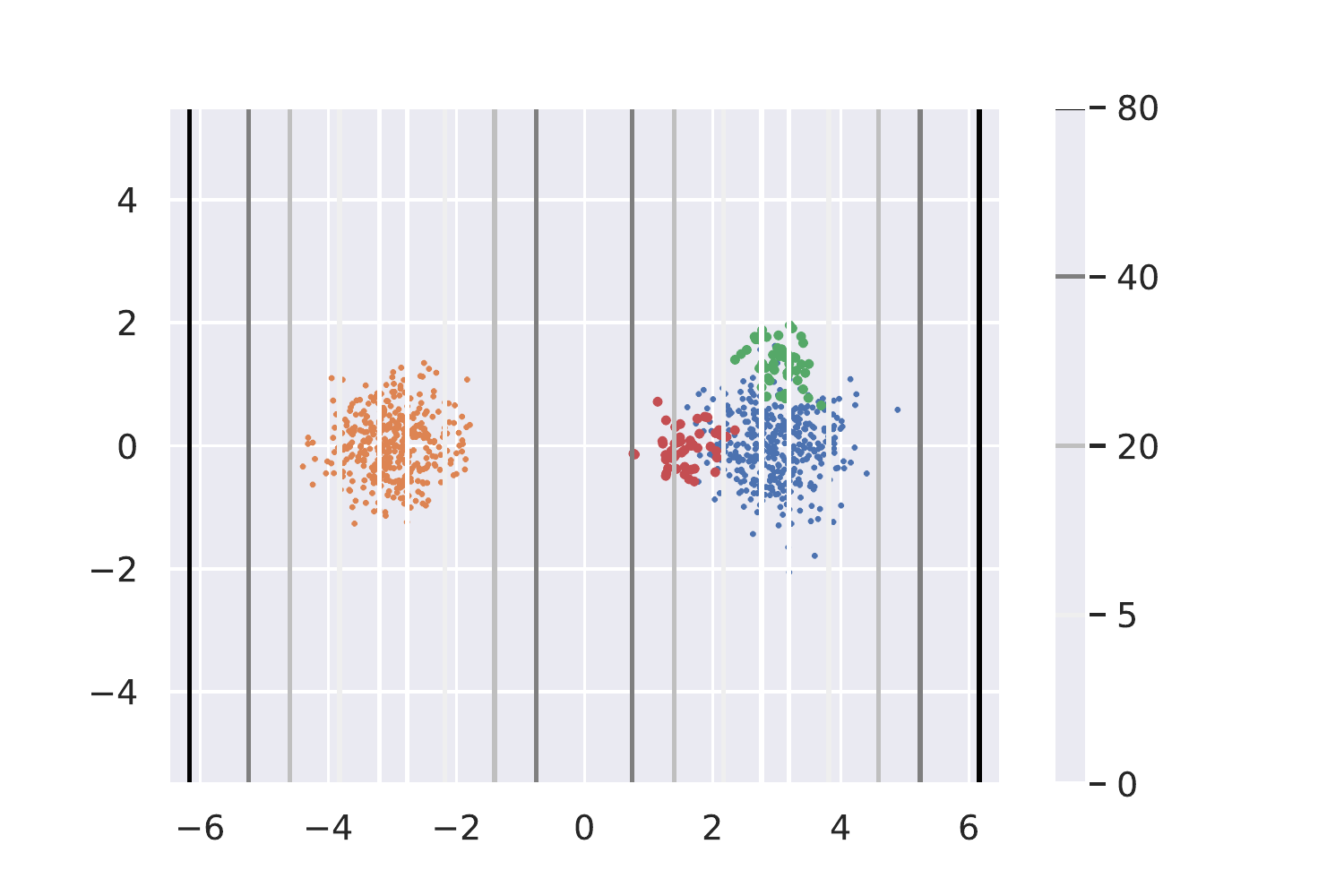}
          \caption{Mahalanodis distance on discriminative features. \newline \textbf{AUROC$_{red}$= 0.98}, \newline AUROC$_{green} = 0.36.$ }
         \label{analytic_dis}
    \end{subfigure}
    \begin{subfigure}{0.325\textwidth}
        \centering
        \includegraphics[width=\textwidth]{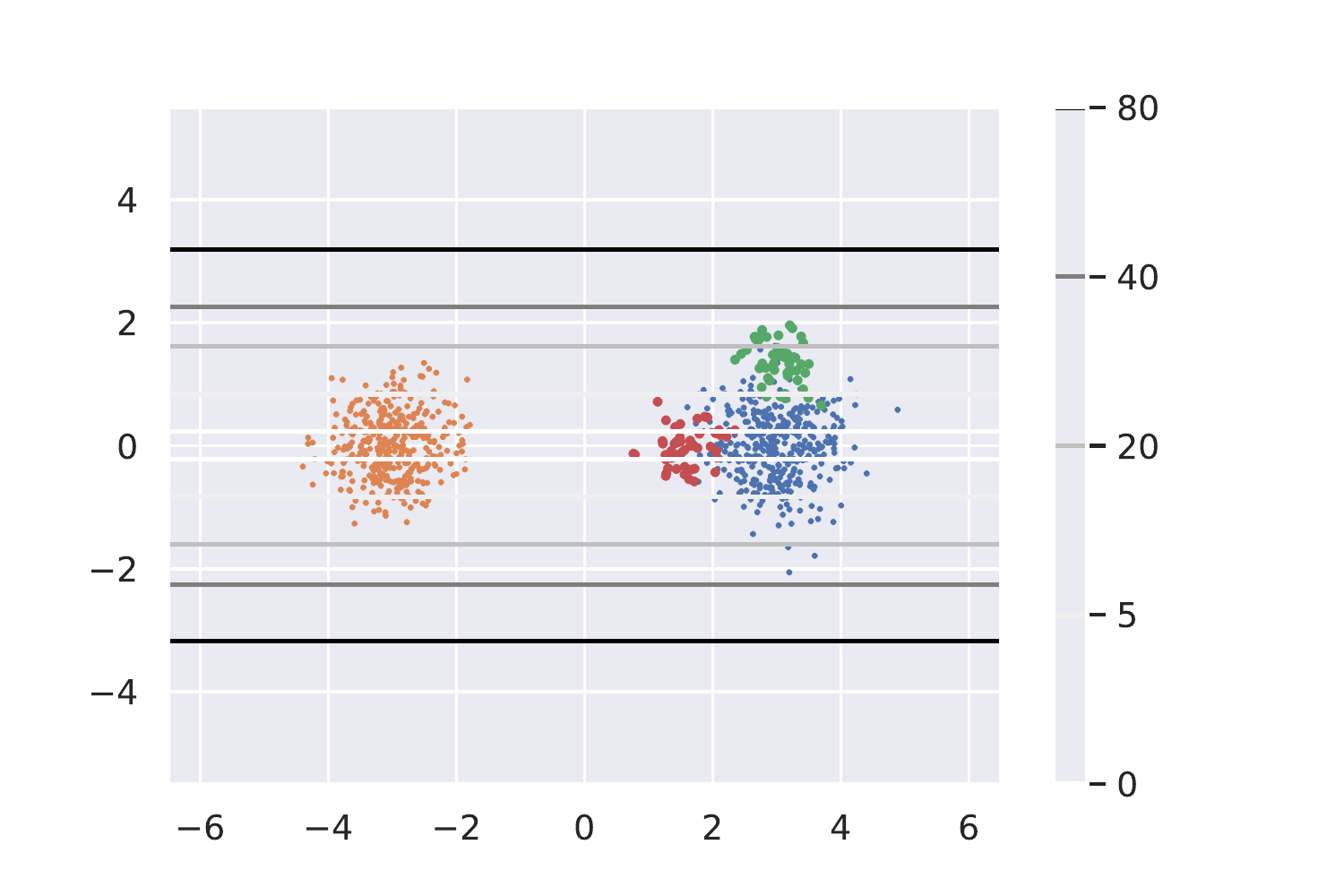}
         \caption{Mahalanodis distance on non-discriminative features. \newline \textbf{AUROC$_{green}$ = 0.98},  \newline AUROC$_{red} = 0.34.$}
        \label{analytic_nondis}
   \end{subfigure}
    \caption{A 2D example showing the effect of decomposition.
    The \textit{orange} and \textit{blue} points are two in-distribution classes, \textit{green} points are OoD from non-discriminative directions and \textit{red} points are OoD from discriminative directions.  
    Contour lines are colored according to the Mahalanobis distance to the in-distribution points (darker means higher).
    We can see that  \textit{Maha} on the full feature weighs on both discriminative and non-discriminative directions simultaneously (and in our case, equally). This makes \textit{Maha} on full features (\textbf{(a)}) suboptimal in detecting OoD data  compared to \textit{Maha} calculated on either direction only (\textbf{(b), (c)}).}
    \label{fig:toy_experiment}
\end{figure}

We consider both low-dimensional (2D) simulation and high-dimensional (128D) simulation. For the low dimensional simulation, we generate two Gaussian distributions with mean (3,0) and (-3,0) and tied covariance matrix $\Sigma = 0.5 I_2$. The two OoD distributions are also Gaussian distributions with mean (3,1.4) and (1.6, 0) and the same covariance matrix $\Sigma = 0.3 I_2$. We then use Equation~\eqref{maha} to calculate the Mahalanobis distance on the chosen features (full, discriminative, non-discriminative). The results in Figure~\ref{fig:toy_experiment} show that \textit{Maha} calculated on the full features can only reach sub-optimal performance (94\% AUROC) compared to on the decomposed features for the corresponding type of OoD data (98\% AUROC). 

For the high-dimensional simulation, we generate ten 128D Gaussian distributions $\{\boldsymbol{z}_c \sim \mathcal{N}(\boldsymbol{\mu}_c, \Sigma)\}_{c=1}^{10}$ with mean $\boldsymbol{\mu}_c= 10 \boldsymbol{e}_c$, diagonal covariance matrix $\Sigma = I_{128}$. Here $\boldsymbol{e}_c$  is the standard basis.
 We also consider two types of OoD: 
(1) OoD along discriminative directions: $\boldsymbol{z} \sim \mathcal{N}(\boldsymbol{\mu}_k, \Sigma)$ where $\boldsymbol{\mu}_k = 5 \boldsymbol{e}_{k}$, $k \in \{1,2,...,10\}$.; (2) OoD along non-discriminative directions: $\boldsymbol{z} \sim \mathcal{N}(\boldsymbol{\mu}_{k'}, \Sigma)$ where $\boldsymbol{\mu}_{k'} = 10\boldsymbol{e}_c + 5\boldsymbol{e}_l$, $c \in \{1,2,...,10\}, 11 \leq l \leq 128$. The remaining process of calculating Mahalanobis distance is the same as the low-dimensional simulation. When using Mahalanobis distance on the full feature, the OoD detection performance is $83.95\%$ AUROC. While we we use Mahalanobis distance on the decomposed features, we can achieve $99.13\%$ AUROC on OoD data along discriminative directions and $84.41\%$ along non-discriminative directions. This shows that the \textit{uncertainty estimation is dominated by the non-discriminative representations}. This is similar to the CIFAR10 vs CIFAR100 benchmark (see Table~\ref{main}).

\subsection{Histogram comparison on image datasets}

\begin{figure}
    \centering
    \begin{subfigure}{0.48\textwidth}
         \centering
         \includegraphics[width=\textwidth]{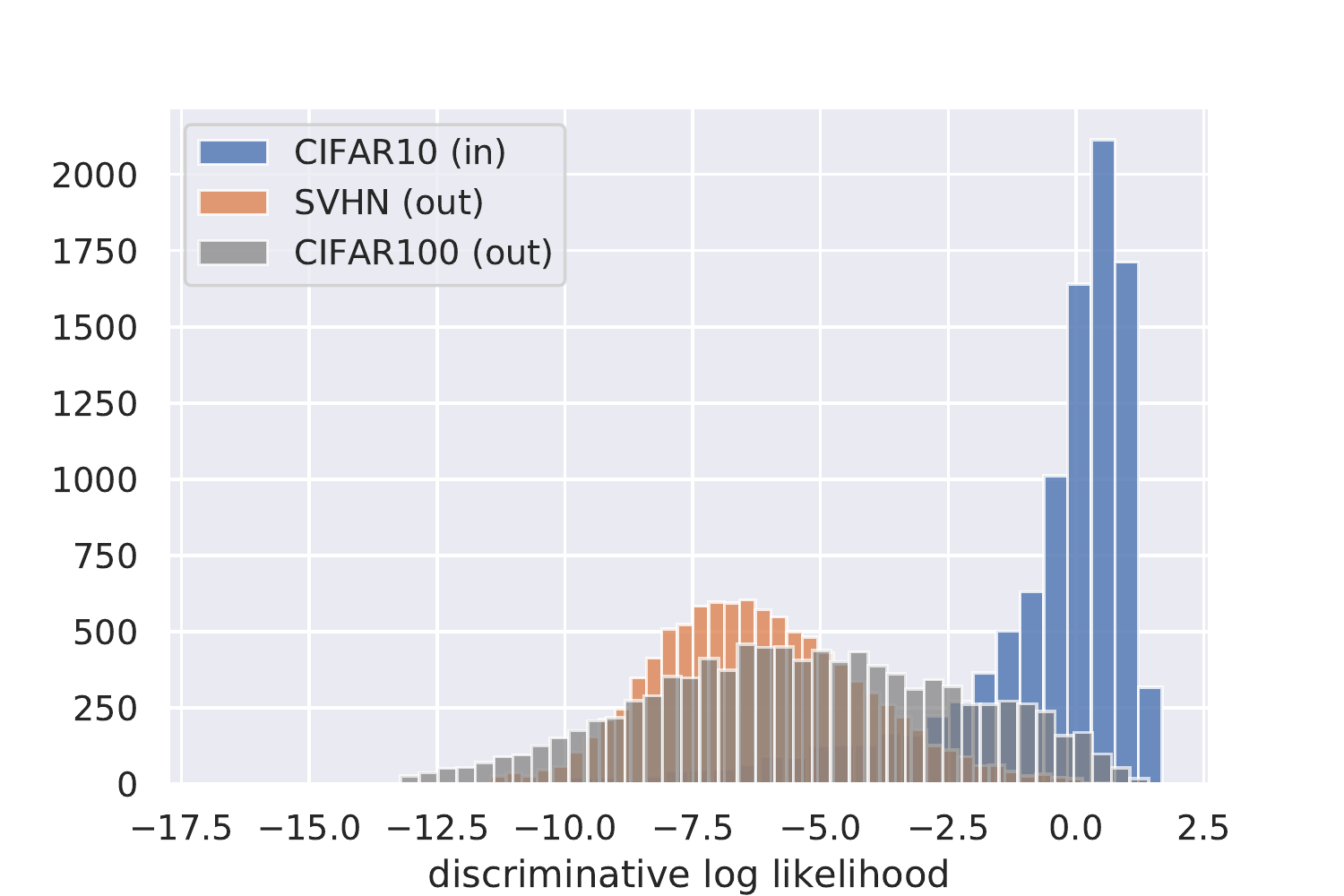}
          \caption{}
         \label{dis_ood}
    \end{subfigure}
    \begin{subfigure}{0.48\textwidth}
         \centering
         \includegraphics[width=\textwidth]{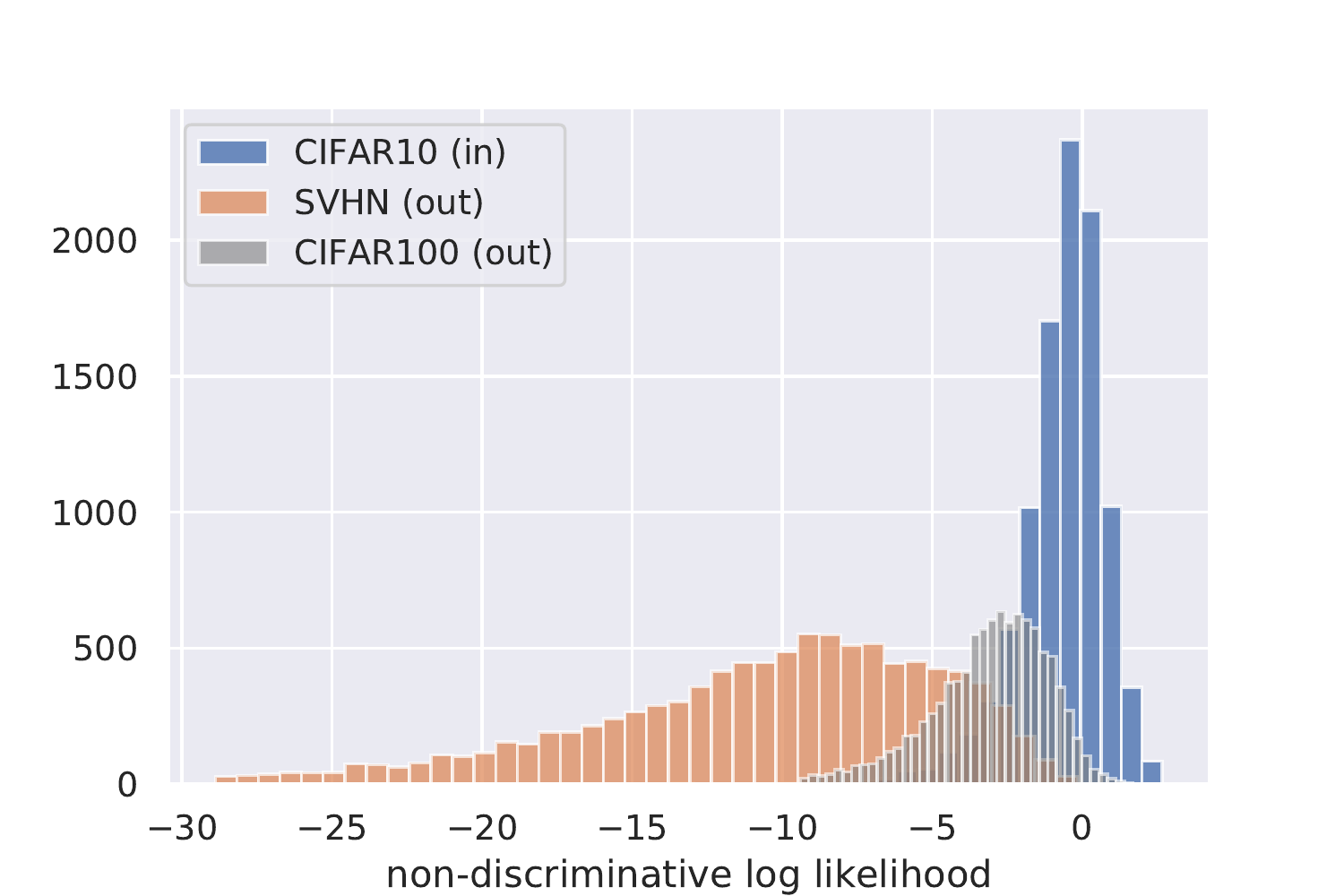}
          \caption{}
         \label{nondis_ood}
    \end{subfigure}
    \caption{OoD detection using uncertainty estimated on different features. CIFAR10 is the in-distribution dataset, and SVHN and CIFAR100 are two different OoD datasets. Notably, when using non-discriminative features, CIFAR100 is much closer to the in-distribution than SVHN. }
    \label{histogram}
\end{figure}

To further understand the dataset distances under two different features, in Figure~\ref{histogram}, we show the histograms of the two scores (log likelihoods) calculated on the three datasets. From the histogram, we can see that  SVHN is far from CIFAR10 in both histograms, but the absolute values of non-discriminative log likelihoods are larger, so they can be easier to detect. For CIFAR100, there is a sharp difference between two features:  the discriminative log likelihoods are more flat and far from CIFAR100, while the non-discriminative log likelihoods are very close to CIFAR10. So discriminative scores are a better choice to detect CIFAR100. These observations are aligned with the AUROC results of OoD detection and the dataset distances in Table~\ref{main}.

\subsection{Setup of experiments on image datasets}
For the experiments on image datasets, we use CIFAR10~\cite{krizhevsky2009learning} as the in-distribution datasets and SVHN~\cite{svhn}, CIFAR100~\cite{krizhevsky2009learning}, and the union of the two datasets as three different benchmarks. This is to demonstrate better how OoD datasets' differences influence the OoD detection performances using different methods.

We use WRN-28-10~\cite{zagoruyko2016wide} trained on CIFAR10 as our pretrained model to extract the penultimate features for different methods. For the decomposition using iCE loss, the invertible transformation is implemented as a 4-layer invertible residual networks~\cite{behrmann_invertible_2019} with linear residual function and 0.9 spectral coefficient. The decomposition takes 3000 iterations using SGD optimizer, and we get 96.12\% test accuracy for discriminative logits and 11.2\% test accuracy for non-discriminative logits.
For each method, we repeat the experiments using five differently randomly initialized models and report the mean.

\section{Discussion of the Decomposition}\label{more_decomposition}
\subsection{Definition of the decomposition}
In Section 3, we listed three properties that an ideal decomposition should possess. For demonstration purposes, we can write them in a slightly redundant way:
\begin{itemize}
    \item $I(z_d; z_n | y) = 0$ (the two decomposed parts should be conditionally independent given the labels $y$);
    \item $I(z_n; y)=0$ ($z_n$ are non-discriminative, i.e., independent from labels $y$);
    \item $I(z_d; y)=I(z; y)$ ($z_d$ includes all the discriminative information in $z$);
    \item $z_d = \argmin_S I(z_d; z)$ where $S = \{z_d \mid I(z_d;y) = I(z;y)\}$ ($z_d$ \textit{only} includes all the discriminative information in $z$).
\end{itemize}
We first show that we can derive the third property from the first two.
To start with, from the first two properties we have  $I(z_d; z_n) = I(z_n; y) + I(z_n; z_d \mid y) - I(z_n;y\mid z_d) = - I(z_n;y\mid z_d) \leq 0$, thus $I(z_d; z_n)=0$. Then since $I(z_d; z_n; y) = I(z_n; y) - I(z_n; y \mid z_d) = I(z_d; z_n) - I(z_d; z_n \mid y)$, we can derive that $I(z_n; y \mid z_d) = I(z_d; z_n; y) = 0$. Finally, combined  with the chain rule of mutual information, $I(z; y) = I(z_d; y) + I(z_n; y \mid z_d)$,  we can derive the third property $I(z;y) = I(z_d; y)$. Similarly, when the first and third properties are satisfied, we can also derive the second property. 

The fourth property is to ensure that $z_d$ does not include more information than discriminative part. For example, simply setting $z_d = z$ and $z_n = 0$ could satisfy the first three properties. 
Note that since $I(z_d ; y) \leq I(z;y)$, to achieve the fourth property, we may use the following optimization scheme:
$$z_d = \argmin I(z_d; z) - I(z_d; y). $$
We can achieve this by adding a cross entropy loss on the logits ($z_d$) based on the information bottleneck method~\cite{tishby2000information}. 

We can also show that when the last three properties are satisfied, the first property will be satisfied as well. 
As a sketch proof, assume the last three properties are already satisfied for $z_d, z_n$, if $I(z_d;z_n|y) > 0$, we can then simply set $z_d'=z_d|z_n$~\footnote{Note $I(z_d;z_n|y) = I(z_d;z_n)$ when the second and third properties hold. This is because $I(z_n;y|z_d) = I(z;y) - I(z_d;y) = 0$, thus $I(z_d; z_n; y) = I(z_n;y) - I(z_n;y|z_d) =0$. And since we also have $I(z_d; z_n; y) = I(z_d; z_n) - I(z_d; z_n|y)$, we now get $I(z_d;z_n|y) = I(z_d;z_n)$. }, then $I(z_d';z) = I(z_d'; z_n) + I(z_d';y) + H(z_d'|y,z_n) = I(z_d;y) + H(z_d|y,z_n) < I(z_d; z_n) + I(z_d;y) + H(z_d|y,z_n) = I(z_d;z)$, while $I(z_n;y)=0$ and $I(z_d';y)=I(z;y)$ still hold. This leads to a contradiction against the last property.
Intuitively, when $z_d$ \textit{only} includes discriminative information and $z_n$ \textit{only} includes the rest information of $z$, $z_d$ and $z_n$ are conditionally independent given labels $y$.


\subsection{Implementation of the decomposition}
The relationship among the properties of the decomposition gives us two options of implementing the decomposition.
One implementation is to maximize $I(z_d;y)$ (and minimize $I(z_d;z)$) and minimize $I(z_n;y)$ simultaneously using iCE loss, as detailed in Section 3 of the paper. From our discussion above, this will also lead to independence of $z_d$ and $z_n$.
Alternatively, we can also put cross entropy loss only on $z_d$ and add an independence regularizer (e.g., HSIC~\cite{hsic}) on $z_d$ and $z_n$. This can enforce the first and third properties in Section 3 (or first, third, and fourth in A.1), and as we demonstrated, the second property will be then satisfied automatically. This can be a future direction for exploration.

It is worth pointing out that throughout our work, we are focusing on the scenario where we are given a pre-trained feature extractor $f$, so $z = f(x)$ is fixed for each $x$. 
In practice, the properties of $z$  also play a big role in the success of the decomposition. 
If the feature extractor $f$ cannot extract semantically meaningful features, the decomposition will also fail.
When the feature extractor is bi-Lipschitz, we will have $I(z;y) = I(x;y)$. In most cases, we can simply think $I(x;y)$ as an upper bound for $I(z_d;y)$ in the decomposition, i.e., we want $z_d$ to extract all the discriminative information in the \textit{input} $x$.

\subsection{Dataset distance metric based on decomposition}\label{distance}
We can design a new dataset distance metric by measuring the difference in the distributions of uncertainty estimates. The decomposition gives such divergence a clear meaning: dataset distances in either discriminative or non-discriminative directions.

To make the scores of the two features comparable, we first normalize the discriminative (dis) and non-discriminative (non-dis) scores of the test data using the mean and standard deviation of the dis and non-dis scores calculated on the training data. 
To reflect the difference between the two distributions of scores $p_{in}( M (z))$ and $p_{out}( M (z))$, we can compute a Kullback–Leibler (KL) divergence $D_{KL}[p_{in}\|p_{out}] = \sum_{k=1}^{K} p_{in, k} \log \frac{p_{in, k}}{p_{out, k}} $. In our case, since we only have the samples drawn from the two distributions $p_{in}, p_{out}$, we use an estimator of the KL-divergence based on k-nearst-neighbor distances~\cite{wang2009divergence}.
When we use the decomposed discriminative and non-discriminative features, we can also define $d_{dis} = D_{KL}[p_{in}(M(z_d))\|p_{out}(M(z_d))]$ and $d_{nondis} = D_{KL}[p_{in}(M(z_n))\|p_{out}(M(z_n))]$.
We show $d_{dis}$ and $d_{nondis}$ in Table~\ref{main}.

\end{document}